\documentclass[10pt,twocolumn,letterpaper]{article}

\usepackage[pagenumbers]{cvpr} 

\usepackage[accsupp]{axessibility}  

\makeatletter
\@namedef{ver@everyshi.sty}{}
\makeatother

\usepackage{acronym}
\usepackage{adjustbox}
\usepackage{booktabs}
\usepackage[font=small,labelfont=bf]{caption}
\usepackage{colortbl}
\usepackage{fmtcount}
\usepackage{forloop}
\usepackage{multirow}
\usepackage{stackengine}
\usepackage{suffix}
\usepackage[para]{footmisc}  
\usepackage{svg}
\usepackage{rotating}
\usepackage{pifont}    
\usepackage{stmaryrd}  
\usepackage{tabularx}
\usepackage[textsize=scriptsize]{todonotes}
\usepackage{xfrac}
\usepackage{xr}
\usepackage{xspace}

\presetkeys{todonotes}{inline}{}  

\makeatletter
\newcommand*{\addFileDependency}[1]{
  \typeout{(#1)}
  \@addtofilelist{#1}
  \IfFileExists{#1}{}{\typeout{No file #1.}}
}
\makeatother

\definecolor{ForestGreen}{HTML}{228b22}  

\DeclareRobustCommand{\nbd}{\nobreakdash-} 

\newcommand{\pnorm}[1]{L\ensuremath{_#1}}

\newcommand{\heading}[1]{\noindent \textbf{\small #1.}}

\newcommand{\addfig}[3][htbp]{%
\begin{figure}[#1]

\centering
\input{Figures/#3}
\label{fig:#3}

\end{figure}
}

\WithSuffix\newcommand\addfig*[3][htbp]{%
\begin{figure*}[#1]

\centering
\input{Figures/#3}
\label{fig:#3}

\end{figure*}
}

\newcommand{\addtbl}[3][htbp]{%
\begin{table}[#1]

\scriptsize
\addtolength{\tabcolsep}{-0.2em}
\renewcommand{\arraystretch}{1.1}
\centering
\input{Tables/#3}
\label{tbl:#3}

\end{table}
}

\WithSuffix\newcommand\addtbl*[3][htbp]{%
\begin{table*}[#1]

\scriptsize
\addtolength{\tabcolsep}{-0.2em}
\renewcommand{\arraystretch}{1.1}
\centering
\input{Tables/#3}
\label{tbl:#3}

\end{table*}
}

\def\eg{\emph{e.g}.~}

\def\vs{\emph{vs}.~}
\def\wrt{w.r.t.~} 

\def\etal{\emph{et al}.\xspace}

\newcommand{\fig}[1]{Figure~\ref{fig:#1}}
\newcommand{\tbl}[1]{Table~\ref{tbl:#1}}

\newcommand{\acromath}[3]{\acrodef{#1}[\(#2\)]{#3}} 
\newcommand{\myac}[1]{\text{\acs{#1}}}  

\newcommand{\shape}[4]{\ensuremath{  
#1 \times #2
\ifthenelse {\equal{#3}{}} {} {\times #3}
\ifthenelse {\equal{#4}{}} {} {\times #4}
}}
\newcommand{\sqwdw}[1]{\shape{#1}{#1}{}{}}  

\newcommand{\sca}[1]{\ensuremath{#1}}  						   
\newcommand{\vct}[1]{\ensuremath{\textbf{\MakeLowercase{#1}}}} 
\newcommand{\mat}[1]{\ensuremath{\textbf{\MakeUppercase{#1}}}} 
\newcommand{\set}[1]{\ensuremath{\MakeUppercase{#1}}} 	       

\newcommand{\thr}{\delta}

\newcommand{\brackets}[3]{\ensuremath{\left#1 #2 \right#3}}

\newcommand{\mybra}[1]{\brackets{\{}{#1}{\}}}

\newcommand{\mysqb}[1]{\brackets{[}{#1}{]}}

\acrodef{ar}     [AR]             {Augmented Reality}
\acrodef{araug}  [AR-Aug]         {aspect ratio augmentation}

\acrodef{cnn}    [CNN]            {Convolutional Neural Network}

\acrodef{d25}    [$\delta_{.25}$] {$\delta < 1.25$}
\acrodef{dnn}    [DNN]            {Deep Neural Network}

\acrodef{fscore} [F]              {F-Score}

\acrodef{imu}    [IMU]            {Inertial Measurement Unit}

\acrodef{kbr}    [KBR]            {Kick Back \& Relax}

\acrodef{lidar}  [LiDAR]          {Light Detection and Ranging}

\acrodef{mae}    [MAE]            {Mean Absolute Error}
\acrodef{mde}    [MDE]            {monocular depth estimation}
\acrodef{mdec}   [MDEC]           {Monocular Depth Estimation Challenge}
\acrodef{ml}     [ML]             {Machine Learning}

\acrodef{rel}    [Rel]            {Absolute Relative Error}

\acrodef{sfm}    [SfM]            {Structure-from-Motion}
\acrodef{slam}   [SLAM]           {Simultaneous Localization and Mapping}
\acrodef{sota}   [SotA]           {State-of-the-Art}
\acrodef{ssl}    [SSL]            {self-supervised learning}

\acrodef{vo}     [VO]             {Visual Odometry}

\acrodef{crf}      [NeWCRFs] {NeWCRFs}

\acrodef{ddad}     [DDAD]    {DDAD}
\acrodef{dii}      [DIODE]   {DIODE Indoors}
\acrodef{dio}      [DIODE]   {DIODE Outdoors}
\acrodef{dpt-beit} [DPT]     {DPT-BEiT}
\acrodef{dpt-vit}  [DPT]     {DPT-ViT}

\acrodef{kb}       [KB]      {Kitti Benchmark}
\acrodef{ke}       [KE]      {Kitti Eigen}
\acrodef{keb}      [KEB]     {Kitti Eigen{\nbd}Benchmark}
\acrodef{kez}      [KEZ]     {Kitti Eigen{\nbd}Zhou}

\acrodef{mc}       [MC]      {Mannequin Challenge}
\acrodef{midas}    [MiDaS]   {MiDaS}

\acrodef{nyud}     [NYUD]    {NYUD{\nbd}v2}

\acrodef{sintel}   [Sintel]  {Sintel}
\acrodef{ssmde}    [SS-MDE]  {SS{\nbd}MDE}
\acrodef{stv}      [STV]     {SlowTV}
\acrodef{syns}     [SYNS]    {SYNS{\nbd}Patches}

\acrodef{tum}      [TUM]     {TUM{\nbd}RGBD}

\acromath{pix}        { \vct{p} } {}

\acromath{point-gt}   { \vct{q} } {}
\acromath{point-pred} { \hat{\myac{point-gt}} } {}
 
\acromath{depth-gt}   { \sca{y} } {}
\acromath{depth-pred} { \hat{\myac{depth-gt}} } {}
 
\acromath{cloud-gt}   { \set{Q} } {}
\acromath{cloud-pred} { \hat{\myac{cloud-gt}} } {}
 
\acromath{edges-gt}   { \mat{M} } {}
\acromath{edges-pred} { \hat{\acs{edges-gt}} } {}
 
\acromath{thr-3d}     { \sca{\thr} } {}

\graphicspath{{Images/}}

\def\challink{\url{https://codalab.lisn.upsaclay.fr/competitions/10031}}

\newcommand{\team}[1]{%
\ifcase#1
\textbf{Baseline}\xspace
\or DJI\&ZJU\xspace
\or Pokemon\xspace
\or cv-challenge\xspace
\or DepthSquad\xspace
\or imec-IDLab-UAntwerp\xspace
\or GMD\xspace
\or MonoViTeam\xspace
\or USTC-IAT-United\xspace
\else
Invalid number%
\fi
}

\newcommand{\train}[1]{%
\ifcase#1
S\xspace
\or D\xspace
\or D\xspace
\or D\xspace
\or D\xspace
\or MS\xspace
\or MS\xspace
\or MSD*\xspace
\or MS\xspace
\else
Invalid number%
\fi
}

\newcommand{\teamsec}[1]{Team #1: \team{#1} -- \train{#1}}

\usepackage[pagebackref,breaklinks,colorlinks]{hyperref}

\usepackage[capitalize]{cleveref}
\crefname{section}{Sec.}{Secs.}
\Crefname{section}{Section}{Sections}
\Crefname{table}{Table}{Tables}
\crefname{table}{Tab.}{Tabs.}

\def\cvprPaperID{1} 
\def\confName{CVPR}
\def\confYear{2023}

\makeatletter
\let\@fnsymbol\@arabic
\makeatother

\begin{document}

\title{The Second Monocular Depth Estimation Challenge}

\author{%
Jaime Spencer\thanks{University of Surrey} \and
C.\ Stella Qian\thanks{Aston University} \and
Michaela Trescakova\thanks{University of Southampton} \and
Chris Russell\thanks{Amazon} \and
Simon Hadfield\footnotemark[1] \and
Erich W.\ Graf\footnotemark[3] \and
Wendy J.\ Adams\footnotemark[3] \and
Andrew J.\ Schofield\footnotemark[2] \and
James Elder\thanks{York University} \and
Richard Bowden\footnotemark[1] \and
Ali Anwar\thanks{imec-University of Antwerp} \and
Hao Chen\thanks{Zhejiang University} \and
Xiaozhi Chen\thanks{DJI Technology} \and
Kai Cheng\thanks{University of Science and Technology of China} \and
Yuchao Dai\thanks{Northwestern Polytechnical University} \and
Huynh Thai Hoa\thanks{DeltaX} \and
Sadat Hossain\footnotemark[11] \and
Jianmian Huang\thanks{Independent} \and
Mohan Jing\footnotemark[9] \and
Bo Li\footnotemark[10] \and
Chao Li\thanks{VIVO} \and
Baojun Li\footnotemark[12] \and
Zhiwen Liu\footnotemark[13] \and
Stefano Mattoccia\thanks{University of Bologna} \and
Siegfried Mercelis\footnotemark[6] \and
Myungwoo Nam\footnotemark[11] \and
Matteo Poggi\footnotemark[14] \and
Xiaohua Qi\footnotemark[9] \and
Jiahui Ren\footnotemark[10] \and
Yang Tang\thanks{East China University of Science and Technology} \and
Fabio Tosi\footnotemark[14] \and
Linh Trinh\footnotemark[6] \and
S.\ M.\ Nadim Uddin\footnotemark[11] \and
Khan Muhammad Umair\footnotemark[11] \and
Kaixuan Wang\footnotemark[8] \and
Yufei Wang\footnotemark[10] \and
Yixing Wang\footnotemark[13] \and
Mochu Xiang\footnotemark[10] \and
Guangkai Xu\footnotemark[9] \and
Wei Yin\footnotemark[8] \and
Jun Yu\footnotemark[9] \and
Qi Zhang\footnotemark[13] \and
Chaoqiang Zhao\footnotemark[15]
}
\maketitle

\begin{abstract}
This paper discusses the results for the second edition of the Monocular Depth Estimation Challenge (MDEC). This edition was open to methods using any form of supervision, including fully-supervised, self-supervised, multi-task or proxy depth. The challenge was based around the SYNS-Patches dataset, which features a wide diversity of environments with high-quality dense ground-truth. This includes complex natural environments, e.g.\ forests or fields, which are greatly underrepresented in current benchmarks. 

The challenge received eight unique submissions that outperformed the provided SotA baseline on any of the pointcloud- or image-based metrics. The top supervised submission improved relative F-Score by 27.62\%, while the top self-supervised improved it by 16.61\%. Supervised submissions generally leveraged large collections of datasets to improve data diversity. Self-supervised submissions instead updated the network architecture and pretrained backbones. These results represent a significant progress in the field, while highlighting avenues for future research, such as reducing interpolation artifacts at depth boundaries, improving self-supervised indoor performance and overall natural image accuracy. 
\end{abstract}

\section{Introduction} \label{sec:intro}
\Ac{mde} refers to the task of predicting the distance from the camera to each image pixel. 
Unlike traditional geometric correspondence and triangulation techniques, this requires only a single image.
Despite the ill-posed nature of the problem, deep learning has shown rapid improvements in this field.

Unfortunately, many existing approaches have focused solely on training and evaluating in an automotive urban setting.
This puts into question their ability to adapt to previously unseen environments.
The proposed \ac{mdec} aims to mitigate this by evaluating models on a complex dataset consisting of natural, agricultural, urban and indoor scenes. 
Furthermore, this is done in a zero-shot fashion, meaning that the models must be capable of generalizing.

The first edition of \ac{mdec}~\cite{Spencer2023} focused on benchmarking self-supervised approaches.
The submissions outperformed the baseline~\cite{Spencer2022,Garg2016} in all image-based metrics (AbsRel, MAE, RMSE), but provided slightly inferior pointcloud reconstructions~\cite{Ornek2022} (F-Score). 
The second edition of \ac{mdec}, detailed in this paper, ran in conjunction with CVPR2023. 
This edition was open to any form of supervision, \eg supervised, self-supervised or multi-task. 
The aim was to evaluate the state of the field as a whole and determine the gap between different supervision strategies.

The challenge was once again centered around \acl{syns}~\cite{Adams2016,Spencer2022}.
This dataset was chosen due its diversity, which includes urban, residential, industrial, agricultural, natural and indoor scenes. 
Furthermore, \acl{syns} contains dense high-quality \acs{lidar} ground-truth, which is exceedingly rare in outdoor environments.
This ensures that the evaluations accurately reflect the capabilities of each model.

Eight teams out of the 28 final submissions outperformed the \ac{sota} baseline in either pointcloud- or image-based metrics. 
Half of these submission were supervised using ground-truth depths, while the remaining half were self-supervised with the photometric reconstruction loss~\cite{Garg2016,Godard2017}.
As expected, supervised submissions typically outperformed self-supervised ones. 
However, the novel self-supervised techniques generally outperformed the provided baseline, even in pointcloud reconstructions. 
The remainder of the paper will provide the technical details of each submission, analyze their results on \acl{syns} and discuss potential directions for future research.

\section{Related Work} \label{sec:lit}

\heading{Supervised}
Eigen~\etal~\cite{Eigen2015} introduced the first end-to-end CNN for \ac{mde}, which made use of a scale-invariant loss and a coarse-to-fine network. 
Further improvements to the network architecture included the use of CRFs~\cite{Liu2015,Weihao2022}, regression forests~\cite{Roy2016}, deeper architectures~\cite{Xian2018,Ranftl2020}, multi-scale prediction fusion~\cite{Miangoleh2021} and transformer-based encoders~\cite{Ranftl2021,Cheng2021,Bhat2023}.
Alternatively, depth estimation was formulated as a discrete classification problem~\cite{Fu2018,Li2019,Bhat2021,Bhat2022}.
In parallel, novel losses were proposed in the form of gradient-based regression~\cite{Li2018,Wang2019b}, the berHu loss~\cite{Laina2016a}, an ordinal relationship loss~\cite{Chen2016} and scale/shift invariance~\cite{Ranftl2020}.

Recent approaches focused on the generalization capabilities of \ac{mde} by training with collections of datasets~\cite{Ummenhofer2017,Facil2019,Ren2019,Ranftl2020,Ranftl2021,Bhat2021}.
This relied on the availability of ground-truth annotations, including automotive data \acs{lidar}~\cite{Geiger2013,Guizilini2020,Huang2019}, RGB-D/Kinect~\cite{Silberman2012,Cho2021,Sturm2012}, SfM reconstructions\cite{Li2018,Li2020}, optical flow/disparity estimation~\cite{Xian2018,Ranftl2020} or crowd-sourced annotations~\cite{Chen2016}. 
These annotations varied in accuracy, which may have impacted the final model's performance. 
Furthermore, this increased the requirements for acquiring data from new sources, making it challenging to scale to larger amounts of data.

\heading{Self-Supervised}
Instead of relying on costly annotations, Garg~\etal~\cite{Garg2016} proposed an algorithm based on view synthesis and the photometric consistency across stereo pairs. 
Monodepth~\cite{Godard2017} incorporated differentiable bilinear interpolation~\cite{Jaderberg2015}, virtual stereo prediction and a SSIM+\pnorm{1} reconstruction loss. 
SfM-Learner~\cite{Zhou2017} required only monocular video supervision by replacing the known stereo transform with a pose estimation network.

Artifacts due to dynamic objects were reduced by incorporating uncertainty~\cite{Klodt2018,Yang2020a,Poggi2020}, motion masks~\cite{Gordon2019,Casser2019,Dai2020}, optical flow~\cite{Yin2018,Ranjan2019,Luo2020} or the minimum reconstruction loss~\cite{Godard2019}.
Meanwhile, robustness to unreliable photometric appearance was improved via feature-based reconstructions~\cite{Zhan2018,Spencer2020,Yu2020} and proxy-depth supervision~\cite{Klodt2018,Rui2018,Watson2019}.
Developments in network architecture design included 3D (un-)packing blocks~\cite{Guizilini2020}, positional encoding~\cite{Bello2021}, transformer-based encoders~\cite{Zhao2022,Agarwal2023}, sub-pixel convolutions~\cite{Pillai2019}, progressive skip connections~\cite{Lyu2021} and self-attention decoders~\cite{Johnston2020,Yan2021,Zhou2021}.

\heading{Challenges \& Benchmarks}
The majority of \ac{mde} approaches have been centered around automotive data. 
This includes popular benchmarks such as Kitti~\cite{Geiger2013,Uhrig2018} or the Dense Depth for Autonomous Driving Challenge~\cite{Guizilini2020}.
The Robust Vision Challenge series~\cite{Zendel2022}, while generalization across multiple datasets, has so far consisted only of automotive~\cite{Geiger2013} and synthetic datasets~\cite{Butler2012,Richter2017}.

More recently, Ignatov~\etal introduced the Mobile AI Challenge~\cite{Ignatov2021}, investigating efficient \ac{mde} on mobile devices in urban settings. 
Finally, the NTIRE2023\cite{Ramirez2023} challenge, concurrent to ours, targeted high-resolution images of specular and non-lambertian surfaces. 

The \acl{mdec} series~\cite{Spencer2023}---the focus of this paper---is based on the MonoDepth Benchmark~\cite{Spencer2022}, which provided fair evaluations and implementations of recent \ac{sota} self-supervised \ac{mde} algorithms.
Our focus lies on zero-shot generalization to a wide diversity of scenes.
This includes common automotive and indoor scenes, but complements it with complex natural, industrial and agricultural environments.

\addtbl*[!t]{syns_cat}{data:syns_cat}

\addfig*[!t]{distrib}{data:distrib}

\addfig*[!t]{syns}{data:syns}

\section{The Monocular Depth Estimation Challenge} \label{sec:meth}
The second edition of the \acl{mdec}\footnote{\challink} was organized on CodaLab~\cite{Pavao2022} as part of a CVPR2023 workshop.
The initial development phase lasted four weeks, using the \acl{syns} validation split.
The leaderboard for this phase was anonymous, where all method scores were publicly available, but usernames remained hidden.
Each participant could see the metrics for their own submission. 

The final challenge stage was open for two weeks.
In this case, the leaderboard was completely private and participants were unable to see their own scores. 
This encouraged evaluation on the validation split rather than the test split.
Combined with the fact that all ground-truth depths were withheld, the possibility of overfitting due to repeated evaluations was severely limited.

This edition of the challenge was extended to any form of supervision, with the objective of providing a more comprehensive overview of the field as a whole.
This allowed us to determine the gap between different techniques and identify avenues for future research.
We report results only for submissions that outperformed the baseline in any pointcloud-/image-based metric on the Overall dataset.

\heading{Dataset}
The challenge is based on the \acl{syns} dataset~\cite{Adams2016,Spencer2022}, chosen due to the diversity of scenes and environments. 
A breakdown of images per category and some representative examples are shown in \tbl{data:syns_cat} and \fig{data:syns}.
\acl{syns} also provides extremely high-quality dense ground-truth \acs{lidar}, with an average coverage of 78.20\% (including sky regions). 
Given the dense ground-truth, depth boundaries were obtained using Canny edge-detection on the log-depth maps. 
This allows us to compute additional fine-grained metrics for these challenging regions.
As outlined in~\cite{Spencer2022}, the images are manually checked to remove dynamic object artifacts. 

\heading{Evaluation}
Participants provided the unscaled disparity prediction for each dataset image. 
The evaluation server bilinearly upsampled the predictions to the target resolution and inverted them into depth maps.
Self-supervised methods trained with stereo pairs and supervised methods using \acs{lidar} or RGB-D data should be capable of predicting metric depth. 
Despite this, in order to ensure comparisons are as fair as possible, the evaluation aligned predictions with the ground-truth using the median depth. 
We set a maximum depth threshold of 100 meters. 

\heading{Metrics}
We follow the metrics used in the first edition of the challenge~\cite{Spencer2023}, categorized as image-/pointcloud-/edge-based.
Image-based metrics represent the most common metrics (MAE, RMSE, AbsRel) computed using pixel-wise comparisons between the predicted and ground-truth depth map. 
Pointcloud-based metrics~\cite{Ornek2022} (F-Score, IoU, Chamfer distance) instead evaluate the reconstructed pointclouds as a whole.
In this challenge, we report reconstruction F-Score as the leaderboard ranking metric. 
Finally, edge-based metrics are computed only at depth boundary pixels. 
This includes image-/pointcloud-based metrics and edge accuracy/completion metrics from IBims-1~\cite{Koch2018}.

\section{Challenge Submissions} \label{sec:submit}
We outline the technical details for each submission, as provided by the authors. 
Each submission is labeled based on the supervision used, including ground-truth (\textbf{D}), proxy ground-truth (\textbf{D*}) and monocular (\textbf{M}) or stereo (\textbf{S}) photometric support frames. 
The first half represent supervised methods, while the remaining half are self-supervised.

\subsection*{Baseline -- S}
\emph{%
\begin{tabular}{ll}
    J.\ Spencer\footnotemark[1] & j.spencermartin@surrey.ac.uk \\
    C.\ Russell\footnotemark[4] & cmruss@amazon.de \\
   S.\ Hadfield\footnotemark[1] & s.hadfield@surrey.ac.uk \\
     R.\ Bowden\footnotemark[1] & r.bowden@surrey.ac.uk \\
\end{tabular}
}

\noindent
Challenge organizers submission from the first edition. 
\\
\heading{Network}
ConvNeXt-B encoder~\cite{Liu2022} with a base Monodepth decoder~\cite{Mayer2016,Godard2017} from~\cite{Spencer2022}.
\\
\heading{Supervision}
Self-supervised with a stereo photometric loss~\cite{Garg2016} and edge-aware disparity smoothness~\cite{Godard2017}.
\\
\heading{Training}
Trained for 30 epochs on \acl{kez} with an image resolution of \shape{192}{640}{}{}.

\subsection*{\teamsec{1}}
\emph{%
\begin{tabular}{ll}
           W.\ Yin\footnotemark[ 8] & yvanwy@outlook.com \\
         K.\ Cheng\footnotemark[ 9] & chengkai21@mail.ustc.edu.cn \\
            G.\ Xu\footnotemark[ 9] & xugk@mail.ustc.edu.cn \\
          H.\ Chen\footnotemark[ 7] & haochen.cad@zju.edu.cn \\
            B.\ Li\footnotemark[10] & libo@nwpu.edu.cn \\
          K.\ Wang\footnotemark[ 8] & wkx1993@gmail.com \\
          X.\ Chen\footnotemark[ 8] & xiaozhi.chen@dji.com \\
\end{tabular}
}

\noindent
\heading{Network} 
ConvNeXt-Large~\cite{Liu2022} encoder, pretrained on ImageNet-22k~\cite{Deng2009}, and a LeReS decoder~\cite{Yin2021} with skip connections and a depth range of $\mysqb{0.3, 150}$ meters. 
\\
\heading{Supervision} 
Supervised using ground-truth depths from a collection of datasets~\cite{Guizilini2020,Houston2021,Yang2019,Cho2021,Chang2019,Cordts2016,Gehrig2021,Antequera2020,Xiao2021,Bauer2019,Zamir2018}.
The final loss is composed of the SILog loss~\cite{Eigen2015}, pairwise normal regression loss~\cite{Yin2021}, virtual normal loss~\cite{Yin2021b} and a random proposal normalization loss (RPNL).
RPNL enhances the local contrast by randomly cropping patches from the predicted/ground-truth depth and applying median absolute deviation normalization~\cite{Singh2020}.
\\
\heading{Training}
The network was trained using a resolution of \shape{512}{1088}{}{}.
In order to train on mixed datasets directly with metric depth, all ground-truth depths were rescaled as 
$
\acs{depth-pred}' = 
\sfrac{\acs{depth-pred} f_c}{f}
,
$
where $f$ is the original focal length and $f_c$ is an arbitrary focal length. 
This way, the network assumed all images were taken by the same pinhole camera, which improved convergence. 
 
\subsection*{\teamsec{2}}
\emph{%
\begin{tabular}{ll}
         M.\ Xiang\footnotemark[10] & xiangmochu@mail.nwpu.edu.cn \\
           J.\ Ren\footnotemark[10] & renjiahui@mail.nwpu.edu.cn \\
          Y.\ Wang\footnotemark[10] & wangyufei777@mail.nwpu.edu.cn \\
           Y.\ Dai\footnotemark[10] & daiyuchao@nwpu.edu.cn \\
\end{tabular}
}

\noindent
\heading{Network}
Two-stage architecture.
The first part was composed of a SwinV2 backbone~\cite{Liu2022b} and a modified NeWCRFs decoder~\cite{Weihao2022} with a larger attention window.
The second stage used an EfficientNet~\cite{Tan2019a} with 5 inputs (RGB, low-res depth and high-res depth) to refine the high-resolution depth.
\\
\heading{Supervision}
Supervised training using \acs{lidar}/synthetic depth and stereo disparities from a collection of datasets~\cite{Huang2019,Baruch2021,Yao2020,Cordts2016,Cho2021,Vasiljevic2019,Yin2020,Yang2019,Hua2020,Xian2020,Roberts2021,Eigen2015,Silberman2012,Hurl2019,Xian2018,Dai2017,Handa2016,Wang2020,Bauer2019,Cabon2020,Wang2019b}.
Losses included the SILog loss~\cite{Eigen2015} ($\lambda = 0.85$) for metric datasets, SILog ($\lambda = 1$) for scale-invariant training, the Huber disparity loss for Kitti disparities and an affine disparity loss~\cite{Ranftl2020} for datasets with affine ambiguities. 
\\
\heading{Training}
The final combination of losses depended on the ground-truth available from each dataset, automatically mixed by learning an uncertainty weight for each dataset~\cite{Kendall2017a}.
Since each dataset contained differently-sized images, they were resized to have a shorter side of 352 and cropped into square patches. 
Some datasets used smaller crops of size \shape{96}{352}{}{}, such that the deepest feature map fell entirely into the self-attention window (\sqwdw{11}).
A fusion process based on \cite{Miangoleh2021} merged low-/high-resolution predictions into a consistent high-resolution prediction. 

\subsection*{\teamsec{3}}
\emph{%
\begin{tabular}{ll}
            C.\ Li\footnotemark[13] & lichao@vivo.com \\
         Q.\ Zhang\footnotemark[13] & zhangqi.aiyj@vivo.com \\
           Z.\ Liu\footnotemark[13] & zhiwen.liu@vivo.com \\
          Y.\ Wang\footnotemark[13] & wangyixing@vivo.com \\
\end{tabular}
}

\noindent
\heading{Network}
Based on ZoeDepth~\cite{Bhat2023} with a BEiT384-L backbone~\cite{Bao2022}.
\\
\heading{Supervision}
Supervised with ground-truth depth from Kitti and NYUD-v2~\cite{Silberman2012} using the SILog loss. 
\\
\heading{Training}
The original ZoeDepth~\cite{Bhat2023} and DPT~\cite{Ranftl2021} were pretrained on a collection of 12 datasets.
The models were then finetuned on Kitti (\shape{384}{768}{}{}) or NYUD-v2 (\shape{384}{512}{}{}) for outdoor/indoor scenes, respectively. 
Different models were deployed on an automatic scene classifier. 
The fine-tuned models were combined with a content-adaptive multi-resolution merging method~\cite{Miangoleh2021}, where patches were combined based on the local depth cue density.
Since the transformer-based backbone explicitly captured long-term structural information, the original double-estimation step was omitted. 

\subsection*{\teamsec{4}}
\emph{%
\begin{tabular}{ll}
           M.\ Nam\footnotemark[11] & mwn0221@deltax.ai \\
       H.\ T.\ Hoa\footnotemark[11] & hoaht@deltax.ai \\
     K.\ M.\ Umair\footnotemark[11] & mumairkhan@deltax.ai \\
       S.\ Hossain\footnotemark[11] & sadat@deltax.ai \\
 S.\ M.\ N.\ Uddin\footnotemark[11] & sayednadim@deltax.ai \\
\end{tabular}
}

\noindent
\heading{Network}
Based on the PixelFormer architecture~\cite{Agarwal2023} which used a Swin~\cite{Liu2021b} encoder and self-attention decoder blocks with cross-attention skip connections.
Disparity was predicted as a discrete volume~\cite{Bhat2021}, with the final depth map given as the weighted average using the bin probabilities.
\\
\heading{Supervision}
Supervised using the SILog loss \wrt the \acs{lidar} ground-truth.
\\
\heading{Training}
The model was trained on the \ac{kez} split using images of size \shape{370}{1224}{}{} for 20 epochs.
Additional augmentation was incorporated in the form of random cropping and rotation, left-right flipping and CutDepth~\cite{Ishii2021}.
When predicting on \acl{syns}, images were zero-padded to \shape{384}{1248}{}{} to ensure the compatibility of the training resolution. 
These borders were remove prior to submission.

\subsection*{\teamsec{5}}
\emph{%
\begin{tabular}{ll}
         L.\ Trinh\footnotemark[ 6] & khaclinh.trinh@student.uantwerpen.be \\
         A.\ Anwar\footnotemark[ 6] & ali.anwar@uantwerpen.be \\
      S.\ Mercelis\footnotemark[ 6] & siegfried.mercelis@uantwerpen.be \\
\end{tabular}
}

\noindent
\heading{Network}
Pretrained ConvNeXt-v2-Huge~\cite{Woo2023} encoder with an HR-Depth decoder~\cite{Lyu2021}, modified with deformable convolutions~\cite{Dai2021}.
The pose network instead used ResNet-18~\cite{He2016}.
\\
\heading{Supervision}
Self-supervised using the photometric loss~\cite{Godard2019} and edge-aware smoothness.
\\
\heading{Training}
Trained on the \ac{keb} split with images of size \shape{192}{640}{}{}.
The network was trained for a maximum of 30 epochs, with the encoder remaining frozen after 6 epochs.

\subsection*{\teamsec{6}}
\emph{%
\begin{tabular}{ll}
            B.\ Li\footnotemark[12] & 1966431208@qq.com \\
         J.\ Huang\footnotemark[12] & huang176368745@gmail.com \\
\end{tabular}
}

\noindent
\heading{Network}
ConvNeXt-XLarge~\cite{Liu2022} backbone and an HR-Depth~\cite{Lyu2021} decoder. 
\\
\heading{Supervision}
Self-supervised based on the photometric loss~\cite{Godard2019}.
\\
\heading{Training}
Trained on \ac{kez} using a resolution of \shape{192}{640}{}{}.

\subsection*{\teamsec{7}}
\emph{%
\begin{tabular}{ll}
          C.\ Zhao\footnotemark[15] & zhaocq@mail.ecust.edu.cn \\
         M.\ Poggi\footnotemark[14] & m.poggi@unibo.it \\
          F.\ Tosi\footnotemark[14] & fabio.tosi5@unibo.it \\
          Y.\ Tang\footnotemark[15] & yangtang@ecust.edu.cn \\
     S.\ Mattoccia\footnotemark[14] & stefano.mattoccia@unibo.it \\
\end{tabular}
}

\noindent
\heading{Network} 
MonoViT~\cite{Zhao2022} architecture, composed of  MPViT~\cite{Youngwan2022} encoder blocks and a self-attention decoder.
\\
\heading{Supervision} 
Self-supervised on \ac{ke} using the photometric loss~\cite{Godard2019} (stereo and monocular support frames) and proxy depth regression.
Regularized using edge-aware disparity smoothness~\cite{Godard2017} and depth gradient consistency \wrt the proxy labels. 
\\
\heading{Training}
Proxy depths were obtained by training a self-supervised RAFT-Stereo network~\cite{Lipson2021} on the trinocular Multiscopic~\cite{Yuan2021} dataset.
The stereo network was trained for 1000 epochs using \shape{256}{480}{}{} crops.
The monocular network was trained on \ac{ke} for 20 epochs using images of size \shape{320}{1024}{}{}.

\subsection*{\teamsec{8}}
\emph{%
\begin{tabular}{ll}
            J.\ Yu\footnotemark[ 9] & harryjun@ustc.edu.cn \\
          M.\ Jing\footnotemark[ 9] & jing\_mohan@mail.ustc.edu.cn \\
            X.\ Qi\footnotemark[ 9] & xiaohua000109@163.com \\
\end{tabular}
}

\noindent
\heading{Network}
Predictions were obtained as a mixture of multiple networks: DiffNet~\cite{Zhou2021}, FeatDepth~\cite{Shu2020} and MonoDEVSNet~\cite{Gurram2021}.
DiffNet and FeatDepth used a ResNet backbone, while MonoDEVSNet used DenseNet~\cite{Huang2019}.
\\
\heading{Supervision}
Self-supervised using the photometric loss~\cite{Godard2019}. 
\\
\heading{Training}
The three models were trained with different resolutions: \shape{320}{1024}{}{}, \shape{376}{1242}{}{}, \shape{384}{1248}{}{}, respectively. 
All predictions were interpolated to \shape{376}{1242}{}{} prior to ensembling using a weighted average with coefficients $\mybra{0.35, 0.3, 0.35}$.

\section{Results} \label{sec:res}
Participant submissions were evaluated on \acl{syns}~\cite{Adams2016,Spencer2022}. 
As previously mentioned, this paper only discusses submissions that outperformed the baseline in any pointcloud-/image-based metric across the Overall dataset.
Since both challenge phases ran independently and participants were responsible for generating the predictions, we cannot guarantee that the testing/validation metrics used the same model. 
We therefore report results only for the test split. 
All methods were median aligned \wrt the ground-truth, regardless of the supervision used. 
This ensures that the evaluations are identical and comparisons are fair. 

{ 
\addtbl*[!t]{results}{res:results}
\newpage \mbox{} \newpage \mbox{}
}

\addfig*[!t]{depth}{res:depth_viz}

\subsection{Quantitative Results}
\tbl{res:results} shows the overall performance for each submission across the whole dataset, as well as each category. 
Each subset is ordered using F-Score performance. 
We additionally show the ranking order based on Overall F-Score for ease of comparison across categories. 

The Overall top F-Score and AbsRel were obtained by Team~\team{1}, supervised using ground-truth depths from a collection of 10 datasets. 
This represents a relative improvement of 27.62\% in F-Score (13.72\% -- Baseline) and 18\% in AbsRel (29.66\% -- OPDAI) \wrt the first edition of the challenge~\cite{Spencer2023}.
The top-performing self-supervised method was Team~\team{5}, which leveraged improved pretrained encoders and deformable decoder convolutions.
This submission provided relative improvements of 16.61\% F-Score and 4.04\% AbsRel over the first edition.

As expected, supervised approaches using ground-truth depth generally outperformed self-supervised approaches based on the photometric error. 
However, it is interesting to note that supervising a model with only automotive data (\eg Team~\team{4}, trained on \ac{kez}) was not sufficient to guarantee generalization to other scene types.
Meanwhile, as discussed in~\cite{Spencer2022}, improving the pretrained backbone (Teams~\team{5} \& \team{6}) is one of the most reliable ways of increasing performance. 
Alternative contributions, such as training with proxy depths (\team{7}) or ensembling different architectures (\team{8}), can improve traditional image-based results but typically result in slightly inferior reconstructions.

The top submission (\team{1}) consistently outperformed the other submissions across each scene category, demonstrating good generalization capabilities.
However, Teams \team{2} \& \team{3} provided slightly better pointcloud reconstructions in Natural scenes. 
We theorize this might be due to the use of additional outdoor datasets, while \team{1} primarily relies on automotive data. 
It is further interesting to note that self-supervised approaches such as Teams~\team{5} \& \team{6} outperformed even some supervised methods in Urban reconstructions, despite training only on Kitti.

Finally, supervised methods provided the largest improvement in Indoor scenes, since self-supervised approaches were limited to urban driving datasets.
\team{1} relied on Taskonomy and DIML, \team{2} on ScanNet, SceneNet, NYUD-v2 and more and \team{3} made use of ZoeDepth~\cite{Bhat2023} pretrained on the DPT dataset collection~\cite{Ranftl2021}. 
This demonstrates the need for more varied training data in order to generalize across multiple scene types. 

\subsection{Qualitative Results}
\fig{res:depth_viz} shows visualizations for each submission's predictions across varied scene categories. 
Generally, all approaches struggle with thin structures, such as the railings in images two and five or the branches in image four. 
Models vary between ignoring these thin objects (Baseline), treating them as solid objects (\team{8}) and producing inconsistent estimates (\team{3}). 
Self-supervised methods are more sensitive to image artifacts (\eg saturation or lens flare in images one and three) due to their reliance on the photometric loss. 
Meanwhile, supervised methods can be trained to be robust to the artifacts as long as the ground-truth is correct. 

Object boundaries still present challenging regions, as demonstrated by the halos produced by most approaches. 
Even Team~\team{1}, while reducing the intensity of these halos, can sometimes produce over-pixelated boundaries.
However, it is worth pointing out that many submissions significantly improve over the Baseline predictions~\cite{Spencer2022}.
In particular, Teams~\team{3}, \team{5} \& \team{6} show much greater levels of detail in Urban and Agricultural scenes, reflected by the improved Edge-Completion metric in \tbl{res:results}.
This is particularly impressive given the self-supervised nature of some of these submissions. 

Unfortunately, self-supervised approaches show significantly inferior performance in Indoor settings, as they lack the data diversity to generalize. 
This can be seen by the fact that many self-supervised approaches produce incorrect scene geometry and instead predict ground-planes akin to outdoors scenes.  

Images six, thirteen and sixteen highlight some interesting complications for monocular depth estimation.
Transparent surfaces, such as the glass, are not captured when using \acs{lidar} or photometric constraints.
As such, most approaches ignore them and instead predict the depth for the objects behind them.
However, as humans, we know that these represent solid surfaces and obstacles that cannot be traversed. 
It is unclear how an accurate supervision signal could be generated for these cases.
This calls for more flexible depth estimation algorithms, perhaps relying on multi-modal distributions and discrete volumes.

\section{Conclusions \& Future Work} \label{sec:conclusion}
This paper has summarized the results for the second edition of \ac{mdec}.
Most submissions provided significant improvements over the challenge baseline. 
Supervised submissions typically focused on increasing the data diversity during training, while self-supervised submissions improved the network architecture. 

As expected, there is still a performance gap between these two styles of supervision. 
This is particularly the case in Indoor environments. 
This motivates the need for additional data sources to train self-supervised models, which are currently only trained on automotive data.
Furthermore, accurate depth boundary prediction is still a highly challenging problem.
Most methods frequently predicted ``halos'', representative of interpolation artifacts between the foreground and background. 

Future challenge editions may introduce additional tracks for metric \vs relative depth prediction, as predicting metric depth is even more challenging. 
We hope this competition will continue to bring researchers into this field and strongly encourage any interested parties to participate in future editions of the challenge.

\subsection*{Acknowledgments}
This work was partially funded by the EPSRC under grant agreements EP/S016317/1, EP/S016368/1, EP/S016260/1, EP/S035761/1.

{
\small
\bibliographystyle{ieee_fullname}
\bibliography{main}
}

\end{document}